\documentclass{article}

\usepackage[utf8]{inputenc} % allow utf-8 input
\usepackage[T1]{fontenc}    % use 8-bit T1 fonts
\usepackage{hyperref}       % hyperlinks
\usepackage{url}            % simple URL typesetting
\usepackage{booktabs}       % professional-quality tables
\usepackage{amsfonts}       % blackboard math symbols
\usepackage{nicefrac}       % compact symbols for 1/2, etc.
\usepackage{microtype}      % microtypography
\usepackage{algpseudocode}
\usepackage{amsmath,amssymb,amsthm}
\usepackage[square,numbers]{natbib}
\usepackage{amsmath,amsfonts,bm}
\usepackage[english]{babel}
\usepackage{mathtools}
\usepackage{subfig}
\usepackage{algorithm}
\usepackage[toc,page]{appendix}
\usepackage[table]{xcolor}
\usepackage{arxiv,times}

\setcitestyle{square,citesep={,},aysep={,},yysep={,}}
\newcommand{\eqdef}{\stackrel{\vartriangle}{=}}

\newcommand{\bz}{\mathbf{z}}

\newcommand{\norm}[1]{\left\lVert#1\right\rVert}
\usepackage{amsmath,amsfonts,bm}

\def\eqref#1{equation~\ref{#1}}
\def\1{\bm{1}}

\DeclareMathAlphabet{\mathsfit}{\encodingdefault}{\sfdefault}{m}{sl}
\SetMathAlphabet{\mathsfit}{bold}{\encodingdefault}{\sfdefault}{bx}{n}

\newcommand{\R}{\mathbb{R}}

\DeclareMathOperator*{\argmin}{arg\,min}

\DeclareMathOperator{\sign}{sign}

\definecolor{green}{rgb}{0.0,0.5,0.1}
\definecolor{blue}{rgb}{0.01, 0.1, 0.7}

\title{MadNet: Using a MAD Optimization for Defending Against Adversarial Attacks}
\author{%
  \textbf{Shai Rozenberg}\\
  Technion – Israel Institute of Technology\\
  shairoz@cs.technion.ac.il
  \and 
  \textbf{Gal Elidan}\\
  Google Research\\
  elidan@google.com
  \and \textbf{Ran El-Yaniv}\\
    Technion\\
 rani@cs.technion.ac.il
  }
  
\date{May 2020}

\begin{document}
\maketitle

\begin{abstract}
This paper is concerned with the defense of deep models against adversarial attacks.
Inspired by the certificate defense approach, we propose a maximal adversarial distortion (MAD) optimization method for robustifying deep networks. MAD captures the idea of increasing separability of class clusters in the embedding space while decreasing the network sensitivity to small distortions.
Given a deep neural network (DNN) for a classification problem, an application of MAD optimization results in MadNet, a version of the original network, now equipped with an adversarial defense mechanism.
MAD optimization is intuitive, effective and scalable, and
the resulting MadNet can improve the original accuracy.
We present an extensive empirical study demonstrating that 
MadNet improves adversarial robustness performance compared to state-of-the-art methods. 
\end{abstract}

\section{Introduction}

    Defending machine learning models from adversarial attacks has become an increasingly pressing issue as deep neural networks (DNNs) are utilized in evermore facets of daily life. 
    Adversarial attacks can  effectively fool deep models and force them to misclassify, using a slight but maliciously-designed distortion
that, in the image domain, is typically invisible to the human eye \cite{szegedy2013intriguing,goodfellow2014explaining,kurakin2016adversarial,carlini2017towards,athalye2018obfuscated}.
    Despite the advances in protecting against such attacks 
    (see Section~\ref{sec:related}), defense mechanisms are still wanting.

\begin{figure}[!t]
\centering
\begin{subfloat}[Embedding centroid proximity. $L_2$ matrix of ResNet-56 trained on CIFAR-10. The high value in entry $(i,j)$ indicates that class $i$ is in close proximity to class $j$ in the embedding space.]{  \label{fig:distances}
        \includegraphics[scale=0.35]{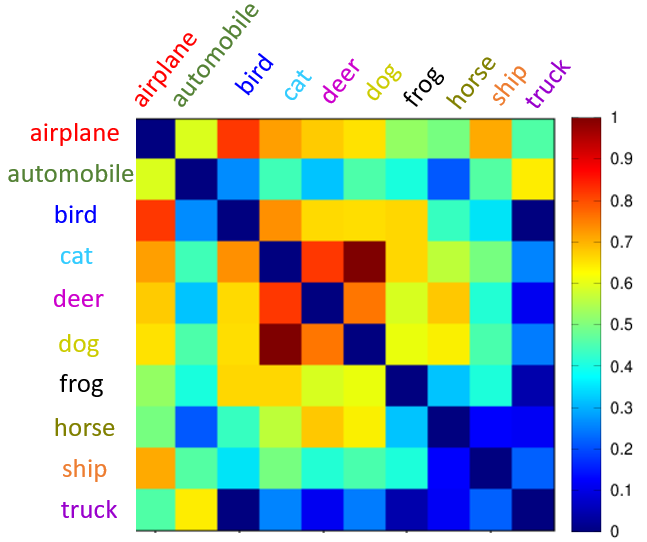}}
\end{subfloat} \hfill
\begin{subfloat}[CIFAR-10 adversarial confusion matrix.  The high value in entry $(i,j)$ represents successful attacks, which change class $i$ to class $j$. ]{\label{fig:confusion}
\includegraphics[scale=0.35]{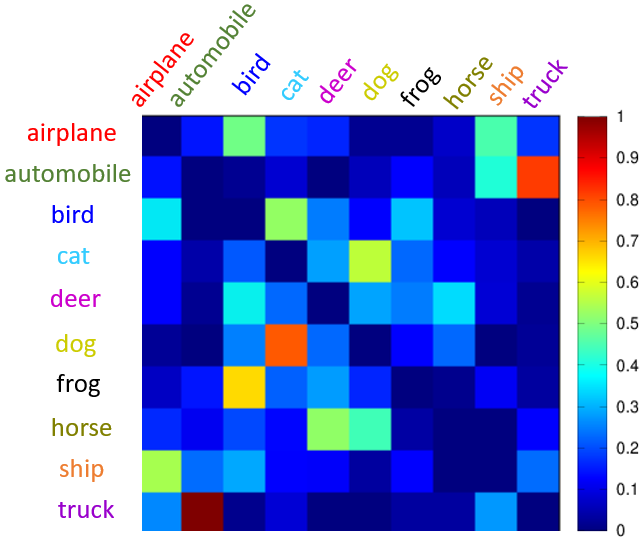}}
\end{subfloat} 
%  \caption{CIFAR-10: C\&W  adversarial confusion matrix and corresponding embedding distance matrix. }
  \caption{CIFAR-10: DNN embedding distance matrix and C\&W  adversarial confusion matrix.}
 \label{fig:both}
\end{figure}

In this paper we study adversarial attacks through the lens of class activation geometry in embedding layers.
We observe
that a trained deep classification model tends to organize instances into clusters in the embedding space,
according to class labels. Classes with clusters in close proximity to  one another
provide excellent opportunities for attackers to fool the model. 
This activation geometry explains the tendency of untargeted attacks to 
alter the label of, for example, a given image to that of an  adjacent class in the embedding space.
This observation is illustrated in Figure~\ref{fig:both} as follows.
We trained Resnet-56 \cite{he2016deep} over the CIFAR-10 \cite{krizhevsky2009learning} image dataset and
calculated the class centroids, computed 
using the $L_2$ mean of class embeddings (pre-logits activation).
Figure~\ref{fig:distances} depicts in colors the distances between class centroids. 
For instance, the closest class to \emph{truck} is \emph{automobile} and the closest class to \emph{dog} is \emph{cat}.
We then attacked this network using the 
C\&W (untargeted) attack \cite{carlini2017adversarial},
and generated many adversarial examples.
Figure~\ref{fig:confusion} depicts the class confusion matrix of this ResNet 
obtained by classifying the
adversarial instances.
We see, for example, that images from class \emph{truck} tend to be adversarialy perturbed so that they are
classified as \emph{automobile}, as indicated by the red matrix cell,
indicating a frequent label change between these classes. 
Similarly, classes \emph{dog} is classified as \emph{cat}, also a frequent misclassification as a result of this attack.
The Euclidean distance provides only partial information about the identity of the adversarial class. 
In untargeted attacks in which the attacker creates an adversarial example with no constraints on the adversarial class, performs a ``step'' in the direction of the gradient. As such, the qualitative nature of the observation.

% As we focus on untargeted attacks in which the attacker creates an adversarial example with no constraints on the adversarial class,
% the attacker creates the adversarial instance by stepping in the direction of the gradient at each layer. The Euclidean distance provides only partial information about the identity of the adversarial class, hence the qualitative observation.

%This corresponds to the two classes embedding clusters being in close proximity. 
% Such cluster geometry observations intuitively suggest that if  
% we modify the model, without compromising accuracy, so as
% to increase the margin between clusters, 
% while lowering (or not increasing) the activation sensitivity in the embedding space to input changes, such a strategy would make the network more immune to attacks.
Such cluster geometry observations intuitively suggest that, by increasing the margin between clusters,  we could make the network more immune to attacks without compromising accuracy.
Indeed manifestations of similar concepts 
have already been considered in the context of
large margin classification and generalization capabilities
\cite{sun2016depth,liang2017soft,sokolic2017robust}.

%One of the main messages of this paper is, however, that such
%separability maximization can only work if we also lower (or at least not increase) the sensitivity of activation in the embedding space to input changes. The embedding activation sensitivity can be quantified through a Lipschitz constant, or directly via the Jacobian of the embedding layer with respect to the input. 
% The benefits of such large margin classification go beyond providing adversarial robustness including facilitating better classification accuracy and better generalization capabilities
% \cite{sun2016depth,liang2017soft,sokolic2017robust}.
%Such large margin classification could potentially have benefits that go beyond providing adversarial robustness including facilitating better classification accuracy and better generalization capabilities
%\cite{sun2016depth,liang2017soft,sokolic2017robust}.

Here we present an adversarial defense method that captures the notion of increased separability in embedding space. While increased separability motivations for 
defense methods appeared elsewhere \cite{mustafa2019adversarial}, the crux in our method is that 
distances must be normalized by the network 
sensitivity in order to be meaningful. Otherwise, one may get a false sense of separability in the embedding space
with a network that travels large distances 
in this space in response to tiny changes in the 
input space. 
This sensitivity can be quantified
through a Lipschitz constant, or directly via the Jacobian of the embedding layer with respect to the input.

Using a distortion $\epsilon$, the adversary tried to compel a DNN $F$ to misclassify $x+\epsilon$, where $x$ is an input. Ideal, we want to maximize the $\epsilon$ needed for this to succeed so that the adversary will fail when using a subtle attack. Using first-order considerations, we
 propose an approximate lower bound on the effective distortion required by the adversary, which motivates a useful strategy for adding a defense mechanism to any architecture.
The first order bound, 
$\epsilon  \geq \eta(F) / ||J_F(x)||$, which is similar to other known bounds \cite{tsuzuku2018lipschitz},
is given in terms of  $\eta(F)$, the ``embedding margin'' of the network $F$, and $J_F(x)$, the Jacobian of $F$ with respect to $x$ (see details in 
Section~\ref{sec:mad}). The embedding margin, for a given intermediate layer, is the minimum distance 
%(under any $p$-norm) 
between two instances belonging to two different classes.
This relation, which we call the  
\emph{maximal adversarial distortion} (MAD) principle,
motivates a MAD optimization technique comprising a loss function and 
a training routine that maximizes this lower bound without attempting to calculate it. 
While a large margin is intuitively a useful property that is 
widely accepted as an inducer of robust classification
\cite{hoffer2015deep,liu2016large,wen2016discriminative},
the above relation emphasizes the principle that the margin should 
be measured in terms of gradient of the model with respect to the input. Otherwise, a large margin would be meaningless when small
changes to the input  correspond to large distances in the embedding space.

To actualize the MAD approach, we introduce two procedures: 
one to increase the embedding margin, and another for reducing the Jacobian. The margin is increased by 
both proxying and penalizing the in-between cluster distance using the
angular distance, and by explicitly penalizing 
within-cluster variance. The Jacobian is handled 
explicitly by penalizing large embedding Jacobian norms.
Both procedures are encapsulated in a single  MAD optimization procedure,
using a Siamese-like training procedure.
We refer to a network trained with the MAD optimization as MadNet.
An extensive empirical study of MadNet shows results under various threat models, in which we consider FGSM, BIM, C\&W and PGD attacks.
Our experimental procedure adheres strictly to the 
comprehensive evaluation desiderata proposed by \cite{carlini2019evaluating}.
The results we present indicate definitively that the proposed defense method substantially improves state-of-the-art detection.

\section{Maximal Adversarial Distortion (MAD)}
\label{sec:mad}

 In this section we develop the MAD principle, whose goal is to improve normalized separability between classes in the embedding space. 
 The MAD principle will be utilized in Section~\ref{sec:madloss} to develop the MAD optimization approach (loss function and training procedure).
Let $F$ be a neural classifier and let $x \in \R^{h \times w}$ be an instance 
%and $z^c_x \in \R^d$, the pre-logits vector activation for sample $x$,
assumed to have the class label $c = c(x)$. 
Let $\epsilon \in \R^{h \times w}$ be a vector representing an adversarial distortion 
for instance $x$ such that the (successful) adversarial instance is $x_{adv} \eqdef x+\epsilon$ whose 
label is different from $c$; namely, $c_{adv} \eqdef F(x_{adv}) \neq c$.
The attacker's goal is to find the smallest distortion $\epsilon$ so that $F$ misclassifies $x$,
% \vspace{-1.5pt}
$$ \left.  \begin{tabular}{l}
    %\displaystyle 
    $\min_\epsilon ||\epsilon||$ \\ \\
    s.t. $F(x+\epsilon) \neq c(x)$ .
    \end{tabular}
    \right. $$
% \vspace{-1.5pt}
For a successful adversarial attack whose distortion is required to be small, in the spirit of \cite{hein2017formal,ding2018max,tsuzuku2018lipschitz,zhang2019recurjac}, we approximate a prediction for $x_{adv}$ using the first-order Taylor approximation,
% \begin{equation}
% \centering
%     F(x_{adv}) = F(x+\epsilon)   \stackrel{|\epsilon|\ll 1}{\approx} F(x) + J_F(x)\epsilon ,
%     \label{eq:taylor_approximation}
% \end{equation}
$$
    F(x_{adv}) = F(x+\epsilon)   \stackrel{|\epsilon|\ll 1}{\approx} F(x) + J_F(x)\epsilon ,
$$
which is applied here for tensor-valued functions with $J_F(x)$ being the Jacobian
of $F$.
The same approximation applies to the output of any intermediate layer $\ell$. 
Denoting by $F_{\ell}(x)$ the output of layer $\ell$, we have,
%\begin{equation}
$$
    F_{\ell}(x_{adv})  \approx  F_{\ell}(x)  +  J_{\ell}(x)\epsilon.
$$
Taking the Forbenius norm, we get,
% \begin{equation}
% \label{eq:taylor}
%  || J_{\ell}(x)\epsilon|| \approx ||F_{\ell}(x) - F_{\ell}(x_{adv})||.
% \end{equation}
$$
|| J_{\ell}(x)\epsilon|| \approx ||F_{\ell}(x) - F_{\ell}(x_{adv})||.
$$

The Frobenius norm is sub-multiplicative (proof can be found in Appendix \ref{appendix_sec:Frobenius}) so that
\begin{equation}
   || J_{\ell}(x)|| ||\epsilon||  \geq || J_{\ell}(x)\epsilon||  \approx ||F_{\ell}(x) - F_{\ell}(x_{adv})||.
%    \eta \leq  || J_l(x)\epsilon|| \leq || J_l(x)|| ||\epsilon||   
    \label{eq:Frobenius_norm_sub-multiplicativity}
\end{equation}
    %\label{eq:layer_output_approx}
%\end{equation}
For layer $\ell$ in $F$, we thus define its \emph{embedding margin}, 
$$
    \eta_{\ell}(F) \eqdef \argmin_{x_1,x_2, c(x_1) \neq c(x_2)}  ||F_{\ell}(x_1) - F_{\ell}(x_2)||,
$$
and we have that
\begin{equation}
\label{eq:applyingMargin}
|| J_{\ell}(x)|| ||\epsilon||  \geq ||F_{\ell}(x) - F_{\ell}(x_{adv})|| \geq \eta_{\ell}(F).
%  || J_{\ell}(x)\epsilon|| \approx ||F_{\ell}(x) - F_{\ell}(x_{adv})|| \geq \eta_{\ell} 
\end{equation}
Combining  (\ref{eq:applyingMargin}) and (\ref{eq:Frobenius_norm_sub-multiplicativity}) 
(and ignoring the Taylor approximation error),
we lower bound the norm of the  distortion $\epsilon$
in terms of the embedding margin and the norm of the Jacobian,
\begin{equation}
||\epsilon|| \geq \frac{\eta_{\ell}(F)}{|| J_{\ell}(x)||} .
\label{eq:distorion_lower_bound}
\end{equation}
While the attacker's goal is to find a small distortion leading to misclassification,
our goal as the defender is to create a resilient model that forces a maximal distortion.
We refer to (\ref{eq:distorion_lower_bound}) as the
\emph{maximal adversarial distortion} (MAD) principle. 
The MAD strategy is thus to 
explicitly maximize the right-hand side of (\ref{eq:distorion_lower_bound}) with respect to the
embedding layer of the model $F$.
To successfully do so, we must increase the embedding margin $\eta_l(F)$ while
decreasing the norm of the Jacobian
$|| J_{\ell}(x)||$.
Thus, the MAD principle 
tells us that the embedding margin must
be measured in terms of gradient units (gradients of the model with respect to the input),
and it is meaningless to enlarge the
embedding margin alone without bounding the Jacobian.

\section{Increasing Resiliency with MAD Optimization}
\label{sec:madloss}
In this section we describe the MAD optimization
technique, which includes the MAD loss function and training routine.
The MAD loss function (presented below) is 
applied to the final embedding layer of a given network $F$. 
%(often referred to as the ``pre-logits layer''). 
We note that, technically, we can also apply the MAD procedure to any other layer in the model (or several layers simultaneously), but defer such explorations to future work. 

To apply the MAD principle (\ref{eq:distorion_lower_bound}), 
need increase the embedding margin 
$\eta_\ell(F)$ and decrease the Jacobian norm
$\norm{J_{\ell}(x)}$.
We achieve the first objective by
adopting ideas from classical cluster analysis.
Specifically, we observe that the increase in 
the embedding margin, $\eta_l(F)$, can be 
accomplished by increasing the distance between clusters
and reducing their variance. 
Let $\bm{\mu}_c = \frac{1}{N_c}\sum_{i=1}^{N_c}{\bz^c_i}$ be the mean of each cluster, where
$N_c$ is the number of samples from class $c$ and $\bz^c_i$ is the embedding vector of sample $i$ from class $c$ , and let $M$ be the number of classes.
We have that,
\begin{equation*}
\begin{split}
\text{Cluster Variance}  \eqdef & 
    \sum_{c=1}^{M} \frac{1}{N_c}\sum_{i=1}^{N_c} \norm{\bz^c_i - \bm{\mu}_c}_2 \\ 
\text{Cluster Distance}  \eqdef &\frac{1}{M} \sum_{c=1}^{M} 
        \frac{1}{M-1}  \sum_{i \neq c}^{M} \norm{\bz_i - \bm{\mu}_c}_2.
\end{split}
\end{equation*}
To increase the margin, we would like to minimize the following objective,
%maximize the cluster distance
%and minimize the cluster variance; hence,
$$
\text{Margin Objective} = \text{Cluster Variance} - \text{Cluster Distance} .
$$
% A straightforward maximization of the cluster distance is problematic because the distance
% is potentially unbounded. We can, however, proxy this quantity using the angular distance between clusters. To this end, we use the cosine 
% similarity and utilize a Siamese training procedure,
% as described in Section~\ref{sec:Siamese}, to maximize the cluster distance.
% The cluster variance is minimized by including a variance
% minimization term
% in the loss function.

Unlike the cluster variance, maximization of the cluster distance is problematic because the distance
is potentially unbounded.
We can, however, proxy this quantity using the angular distance between clusters, which is bounded. To this end, we use the cosine 
similarity and utilize a Siamese training procedure,
as described in Section~\ref{sec:Siamese}.

% \subsection{Siamese Training}
\subsection{MAD Optimization Components}
\label{sec:Siamese}
To explicitly increase the embedding margin, we propose using \textbf{Siamese training} as follows.
We create a Siamese network  \cite{bromley1994signature}, where each subnetwork is our classifier $F$.
The Siamese network receives two input instances denoted by $x_i^c , x_j^{\Tilde{c}}$,  and generates three outputs: two classification outputs and one auxiliary output for the cosine similarity between each subnetwork's embedding. 
We introduce an additional loss term to force embeddings from different class samples to have a cosine similarity of 0 (and 1 otherwise).
Formally,
$$
\text{SiameseLoss} \eqdef \frac{\bz_i^c \cdot \bz_j^{\Tilde{c}}}{\norm{\bz_i^c} \norm{\bz_j^{\Tilde{c}}}} %\stackrel{!}{=} 
=
\left \{     \begin{tabular}{ccc}
    1,  & if $c=\Tilde{c}$ ; \\
    0,  & else .
    \end{tabular}
    \right.
$$
% \subsection{Reduce Variance Loss}

Inspired by \cite{szegedy2016rethinking}, we include an additional loss term that penalizes each class cluster 
individually for large variance. We refer to this component as the \textbf{``reduce variance loss''} (RVL). 
Formally,

\begin{equation}
\sigma_c \eqdef \frac{1}{N_c}\sum_{i=1}^{N_c}{\norm{\bz^c_i - \bm{\mu}_c}_2} , \ \  \ \ \ \ 
 \text{RVL} \eqdef \frac{1}{N_c}\sum_{c=1}^{N_{classes}} \sigma_c .  
 \label{eq:RVL}
\end{equation}

The variance is estimated per class on each mini-batch, then averaged and minimized as part of the learning process.
%

% \subsection{Explicit Jacobian Reduction}
Finally, we add a \textbf{Jacobian reduction loss}, which explicitly evaluates the Jacobian per mini-batch, and include its norm in the DNN's loss function.
%as done in \citet{jakubovitz2018improving} and shown to improve adversarial robustness. 
Formally, $J_{\ell}(x) = \frac{\partial F_{\ell}(x)}{\partial x}$, where the derivative is
taken for each input in a mini-batch.
To minimize the norm of the Jacobian, we include it in the MAD loss function. 

\subsection{Training with MAD Optimization}

A simultaneous application of all components described above is
obtained using the following MAD loss function,
denoted ${\cal L}_{MAD}$,
% \vspace{-3pt}
 \begin{alignat}{2}
 \label{eq:mad}
 \nonumber
 {\cal L}_{MAD}(x_1,x_2,y_1,y_2 ; F) & \eqdef
 \lambda_{\text{CE}} CE(x_1,y_1;F)
 + \nonumber  \lambda_{\text{CE}}CE(x_2,y_2;F)   \nonumber\\ 
 &+ \nonumber 
 \lambda_{\text{Siam}}
 \left(
    \delta_{y_1,y_2}-\frac{\bz_1\bz_2}{\norm{\bz_1} \norm{\bz_2}}
    \right)^2 \\
&+ \nonumber 
\lambda_{\text{RVL}}\frac{1}{N_c}\sum_{c=1}^{N_{classes}} \sigma_c(x_1) 
+ \nonumber 
\lambda_{\text{RVL}}\frac{1}{N_c}\sum_{c=1}^{N_{classes}} \sigma_c(x_2) \\
&+ \nonumber 
\lambda_{\text{Jacob}}
\norm{
    \frac{\partial F(x_1)}{\partial x_1}
    }  
+ \nonumber 
\lambda_{\text{Jacob}}
\norm{
    \frac{\partial F(x_2)}{\partial x_2}
    } ,
\nonumber
 \end{alignat}
 where $CE(x,y;F)$ is the cross-entropy loss of $x$ given its label $y$,
and  $\lambda_{\text{CE}}, \lambda_{\text{RVL}}, \lambda_{\text{Siam}}$, and $\lambda_{\text{Jacob}}$
 are hyperparameters (taken in our experiments to be $1,1,1,0.01$, respectively).\footnote{ $\lambda_{\text{Jacob}} = 0.01$ was chosen to make
 this 
 %loss 
 component roughly of the 
 same 
 %order of 
 magnitude as the other components.}
In addition, we require a specialized mini-batch construction procedure.
A pseudo-code of the training procedure is given in Algorithm~\ref{alg:psudo} in Appendix \ref{appendix_sec:mad}.
The code is self-explanatory for the most part. 
We note, however, that
an epoch begins by creating a Siamese counterpart for each instance--label pair in a given batch.
With probability $Q$, the Siamese sample is selected from the same class, and its cosine similarity label is set to 1. Otherwise (probability $1 - Q$), the Siamese sample is selected from a different class, and its cosine similarity label is set to 0. Both target classes' activation is set to $\alpha$, a label smoothing hyperparameter (see Appendix \ref{appendix_sec:smoothing}).
Notice that the 
Siamese, RVL and Jacobian norm (Equation~\ref{eq:RVL}) components 
%of the loss function
are computed from the embedding vectors of each mini-batch.

% \section{Threat Models and Adversarial Attacks}
% Rigorous analyses of adversarial defense systems 
% require precise specification of what the adversary knows and can do.
% Such specifications are called \emph{threat models} \cite{kurakin2016adversarial}. 
% To evaluate a defense algorithm's efficiency against adversarial attacks, 
% we consider two threat models, differentiated by the knowledge 
% the adversary has of the classifier (target network) and the defense mechanism being used.
% %\begin{itemize}
%   %\item
  
%   \noindent
%   \textbf{Black-box}: 
%   In this scenario, the adversary has no knowledge of our system.
%   It does not know the target network architecture, cannot access its gradients and does not 
%   know the defense methods. 
%   The adversary can, however, sample the target network for input-output pairs and has access to the dataset used to train the target network.
%   %\item
  
%   \noindent
%   \textbf{White-box}: In this scenario, we assume the adversary has full knowledge of the entire system to be attacked, and can access its gradients to produce the adversarial example. 
% %\end{itemize}
% The specifications of the parameters enclosed in our threat models are detailed in Section \ref{sec:experiments}

\section{Related Work}
\label{sec:related}
Many interesting ideas have been proposed to construct
defense mechanisms against adversarial examples. These can be broadly divided into two categories. 
The first are \textit{active} algorithms that process the input or output of the DNN at prediction time, removing the effects of the adversarial algorithm. 
Amongst these algorithms are: Neural Fingerprinting \cite{dathathri2018detecting}, that transforms the DNN's input such that a specific, pre-defined prediction is expected;
Stochastic Adversarial Pruning \cite{dhillon2018stochastic}, that adds a randomized, dropout-like mechanism to the prediction process; and others.
The second category, to which the algorithm presented in this paper belongs, is that of \textit{passive} defense algorithms. Such algorithms alter the DNN's training process, loss function or architecture to create a network that is more robust to adversarial attacks. One popular passive approach is adversarial training \cite{goodfellow2014explaining,madry2017towards,yan2018deep}, where adversarial examples are introduced during the training procedure, making the DNN more robust to perturbations in the input instance. Another approach is the ensemble approach \cite{tramer2017ensemble,strauss2017ensemble,pang2019improving} where, instead of training a single predictor, the training process includes several different DNNs, all trained together with a single loss function that collects the predictions.

A formal approach to adversarial defense is the
\emph{certification approach} \cite{hein2017formal}, which is designed to provide a lower bound for the penetration distortion attempting to fool a given network.
Certified defense methods are referred to as being either ``exact'' or ``conservative''. In exact methods, no distortion smaller than the certification bound can penetrate or confuse the DNN \cite{hein2017formal,wong2017provable,wong2018scaling,cohen2019certified}.
In conservative methods, the bound is merely a relative metric for comparing DNN robustness to adversarial examples
\cite{ding2018max,tsuzuku2018lipschitz,zhang2019recurjac}.
Both exact and conservative methods have been criticized for being computationally expensive and unscalable. \cite{tjeng2018evaluating,cohen2019certified}. Moreover, so far these techniques have lacked in performance.
In contrast, our pragmatic approach that relies on a lower bound approximation, does not provide a theoretical guarantee but leads to excellent performance.

%By increasing the margin between these clusters without
%increasing the norm of this layer's Jacobian, we make it harder
%for an adversary to alter the label using distortions of the same magnitude

Improving the separability of the class clusters in the intermediate layers of a DNN as a mean of defending against adversarial attack is a notion established by \citet{hein2017formal}. 
Following their work, many defense methods have been suggested to increase the margin between classes while maintaining a feasible training process so as to be applicable to large-scale DNNs. 
\citet{liu2016large} and \citet{wang2018cosface} added  an angular constraint to the loss function, forcing a large angular distance between classes.
\citet{elsayed2018large} suggested a margin increasing loss function that explicitly aims to increase the distance between class clusters, truncating the maximal distance given that increasing the distance is an unbounded term. 
\citet{mustafa2019adversarial} suggested arranging the features of each class in specific convex polytopes. To that end, they employed an auxiliary loss function based on the distance between instance features and a \textit{p}-norm ball around the class centroid. \citet{chan2019jacobian} presented the Jacobian Adversarially Regularized Networks (JARN) method in which they imposed constraints on the DNN's Jacobian and showed their impact on the DNN's robustness to adversarial attacks.
In contrast, our MAD principle (Section~\ref{sec:mad}) states that separability and Jacobian must be optimized together so that separability is gradient-normalized, and it is sub-optimal to consider each of these quantities individually.

\section{Experimental Evaluation}
\label{sec:experiments}
We now evalate the merit of our MadNet approach as a defence mechanism against adversarial attacks. We start by describing the experimental setup, including the precise specification of what the adversary knows and can do. We then present qualitative insight on the embedding separability of MadNets, followed by detailed quantitative evaluation in the face of varied attacks.

\subsection{Experimental Setup}
Following \cite{mustafa2019adversarial,pang2019improving,chan2019jacobian,elsayed2018large}, we evaluated MadNet on three image datasets: MNIST \cite{lecun1998gradient}, CIFAR-10 and CIFAR-100 \cite{krizhevsky2009learning}.
We tested four different attacks: \textit{FGSM, BIM, PGD and C\&W}. For PGD and BIM we used 10 iterations and several values of the maximal $L_\infty$ perturbation $\epsilon$, in line with the 
perturbation used by the baseline \cite{mustafa2019adversarial}. A detailed description of the different attacks can be found in Appendix \ref{appendix_sec:adversarial_attacks}.
We chose ResNet-56 \cite{he2016deep} as the backbone architechture of MadNet. A full description of the training parameters can be found in Appendix \ref{appendix_sec:classifier_params}.
We compare our results to the vanilla ResNet-56 baseline trained with cross-entropy, denoted as \textit{CE}, and to robustness results presented by \citet{mustafa2019adversarial}. In their work, they conducted a thorough ablation study and were able to surpass \cite{madry2017towards,yu2018interpreting,kurakin2016adversarial2,ross2018improving}.
As such, we consider their work as the current state-of-the-art, under a similar threat model and test for the attacks and under the same parameter setting as in their ablation study. For the C\&W attack we used 1000 iterations with the confidence set to 0 and various values of the initial constant $c$.
We also apply a maximal $L_\infty$ perturbation as done by \citet{mustafa2019adversarial}, setting the maximal perturbation to $0.3$ for MNIST and $0.03$ for CIFAR-10 and CIFAR-100.
Results are reported as the accuracy, in percentage,  on 1000 adversarially perturbed images, taken from the test set, which were correctly classified prior to the perturbation.

To evaluate a defense algorithm's efficiency against these adversarial attacks, 
we consider two threat models, differentiated by the knowledge 
the adversary has of the classifier (target network) and the defense mechanism being used.
%\begin{itemize}
  %\item
  
  \noindent
  \textbf{Black-Box Threat Model}: 
  In this model, the adversary has no knowledge of our system.
  It does not know the target network architecture, cannot access its gradients and does not 
  know the defense methods. 
  The adversary can, however, sample the target network for input-output pairs and has access to the dataset used to train the target network.
  %\item
  
  \noindent
  \textbf{White-Box Threat Model}: In this model, we assume the adversary has full knowledge of the entire system to be attacked, and can access its gradients to produce the adversarial example. 
%\end{itemize}
We tested the black-box and white-box threat models on each dataset.

\begin{figure}[t]
\begin{subfloat}[CE; DBI $=$0.98]{  \label{fig:t-SNE_sub_baseline}
        \includegraphics[scale=0.253]{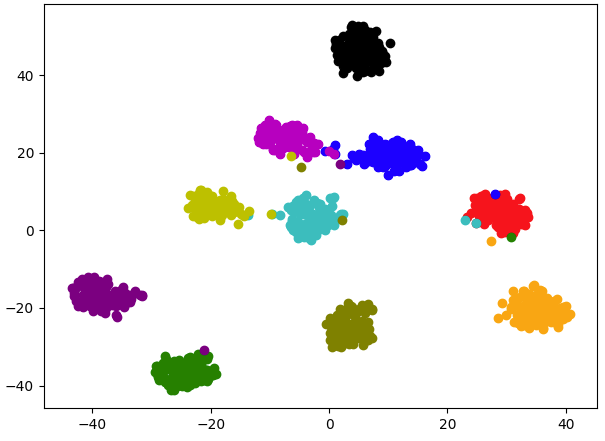}}
\end{subfloat}
% \begin{subfloat}[RCE ; DBI $=$ 0.31]{  \label{fig:t-SNE_sub_rce}
%         \includegraphics[scale=0.3]{images/tsne/vanilla_rce.png}}
% \end{subfloat}
% \begin{subfloat}[Siamese Training; DBI $=$ 0.32]{  \label{fig:t-SNE_sub_siamese}
        % \includegraphics[scale=0.2]{images/tsne/siamese.png}}
% 
        % \includegraphics[scale=0.3]{images/tsne/Mad_tsne_plot.png}}
% \end{subfloat}
% \begin{subfloat}[Siamese $+$ RVL; DBI $=$ 0.28]{  \label{fig:t-SNE_sub_reduce_variance}
        % \includegraphics[scale=0.3]{images/tsne/siamese.png}}
                % \includegraphics[scale=0.2]{images/tsne/reduce_variance.png}}
% \end{subfloat} 
\begin{subfloat}[ MadNet; \textbf{DBI $=$ 0.1}]{  \label{fig:t-SNE_sub_combination}
        \includegraphics[scale=0.27]{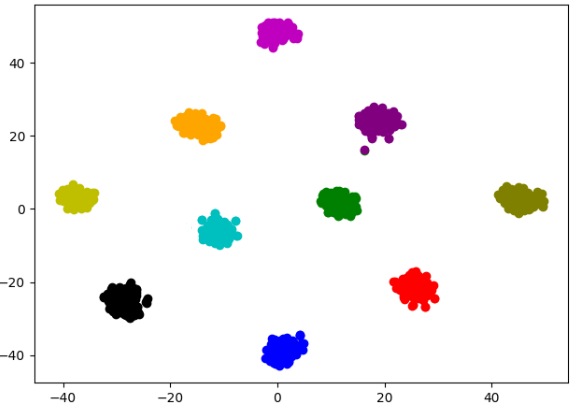}}
\end{subfloat}
\begin{subfloat}{  \label{fig:t-Legend}
\includegraphics[scale=0.45]{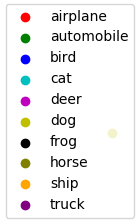}}
\end{subfloat}
\caption{ CIFAR-10 t-SNE and  Davies–Bouldin Index (DBI) for CE and MadNet. }
%visualization  of the two margin increasing components of MadNet. Compared to the baseline, each method contributes to the increase of the margin; MadNet displays the best clustering according to the Davies–Bouldin Index.}
\label{fig:t-SNE_method_overview}
\end{figure}

\subsection{Qualitative Observations on Embedding Separability }
Using t-SNE \cite{maaten2008visualizing} to visualize the embedding space 
activations,
Figure~\ref{fig:t-SNE_method_overview} illustrates how MAD optimization
affects the separability of class embedding clusters.
Figure~\ref{fig:t-SNE_sub_baseline} 
depicts class clusters in the embedding 
layer after standard training over  CIFAR-10 (without MAD).
Figure~\ref{fig:t-SNE_sub_combination} shows
the clusters obtained when using a complete MAD optimization, which also includes Jacobian reduction.
We note that t-SNE is mainly 
useful for visualization and can often distort 
the overall relationships in high dimensions.
To obtain a reliable ablation study, we calculated the Davies–Bouldin index (DBI) \cite{davies1979cluster},
which is frequently used to evaluate clustering quality according to the  distance between cluster centroids divided by the Euclidean distance between points within a cluster (a lower score means better clustering).
DBI scores are shown for all four embedding clusterings in  Figure~\ref{fig:t-SNE_method_overview}.
It is evident that each MAD optimization component contributes significantly to separability,
and the DBI of a complete MAD optimization
is the best by far.

\subsection{White-Box Study}
For the white-box threat model we consider MadNet with two training procedures. The first is where only normal samples are introduced during the training process.
The second is adversarial training where we use the PGD attack during the training of MadNet, and introduce adversarial examples in the training process.
Results for the white-box threat model appear in Table \ref{tab:mnist_white-box}, Table \ref{tab:cifar10_white-box}, and Table \ref{tab:cifar100_white-box}, with the best method without adversarial training highlighted  in \textcolor{green}{green}, and the best method with adversarial training in \textcolor{blue}{blue}. 
Examining these tables, we first note that MadNet does not significantly degrade the performance on normal instances and even improves it for CIFAR-10
(first row in the table). 
It is evident that
MadNet outperforms the competition on the vast majority of the scenarios tested, especially when using adversarial training, improving the results even for attacks not introduced during training.
Moreover, MadNet's performance on MNIST with adversarial training displays near perfect results, creating an almost impenetrable model.
MadNet's performance on CIFAR-10 and CIFAR-100 indicates that it is appropriate for more complex images than MNIST as well as for a larger number of classes.
We note that the significantly better performance for higher-constant C\&W is partially caused by the $L_{\infty}$ bound imposed.
When the $L_{\infty}$ constraint is removed, MadNet achieves an accuracy of 71.6\% with adversarial training and 67.2\% without for MNIST with $c=10$, and 61.5\% with adversarial training for CIFAR-10 with $c=0.1$. 
Other C\&W scenarios were not significantly affected by this constraint.

\begin{table}[!h]
\centering

\caption{MNIST adversarial robustness ($\%$) under the white-box threat model.}
\label{tab:mnist_white-box}
\begin{tabular}{cccccc}
\hline
Attack       & CE  & Mustafa et al. & MadNet    &\begin{tabular}[c]{@{}c@{}}Mustafa et al.\\ Adv training\end{tabular} & \begin{tabular}[c]{@{}c@{}} MadNet \\ Adv training\end{tabular} \\ \hline
No Attack    & 99.22         & 99.5          & 99.3      & 99.3                                                                    & 99.3                                                                       \\
C\&W ($c=0.1$) & 31.6          & 97.7          & \cellcolor{green!25}98.3 & 97.6                                                                    & \cellcolor{blue!25}99.4                                                                  \\
C\&W ($c=1$)   & 0             & 80.4          & \cellcolor{green!25}93.1 & 91.2                                                                    & \cellcolor{blue!25}98.1                                                                  \\
C\&W ($c=10$)  & 0             &29.1          &  \cellcolor{green!25}92.5 & 46.0                                                                    & \cellcolor{blue!25}97.9                                                                  \\
FGSM($\epsilon=0.1$)  & 74            & \cellcolor{green!25}97.1          & 95.3      & 96.5                                                                    &\cellcolor{blue!25} 98.3                                                                       \\
FGSM($\epsilon=0.2$)  & 23            & 70.6          & \cellcolor{green!25}73.5      & 77.9                                                                    & \cellcolor{blue!25}95.8                                                                       \\
BIM($\epsilon=0.1$)   & 17.4          & 90.2          & \cellcolor{green!25}92.2      & 92.1                                                                    & \cellcolor{blue!25}98.6                                                                       \\
BIM($\epsilon=0.15$)  & 3.4           & 70.6          & \cellcolor{green!25}87        & 77.3                                                                    & \cellcolor{blue!25}95.2                                                                       \\
PGD ($\epsilon=0.1$)  & 10.6          & 83.6          & \cellcolor{green!25}92.4      & 93.9                                                                    & \cellcolor{blue!25}98.8                                                                       \\
PGD ($\epsilon=0.15$) & 0.6           & 62.5          & \cellcolor{green!25}83.4      & 80.2                                                                    & \cellcolor{blue!25}98.4               \\ \hline                                                      
\end{tabular}
\end{table}

\begin{table}[!h]
\centering
\caption{CIFAR-10 adversarial robustness ($\%$) under the white-box threat model.}
\label{tab:cifar10_white-box}
\begin{tabular}{cccccc }
\hline
Attack       & CE  & Mustafa et al. & MadNet    &\begin{tabular}[c]{@{}c@{}}Mustafa et al.\\ Adv training\end{tabular} & \begin{tabular}[c]{@{}c@{}} MadNet \\ Adv training\end{tabular} \\ \hline
No-Attack     & 93            & 90.5          & 94.2      & 91.9                                                                       & 92.1        \\
CW ($c=0.001$)  & 30.5          & \cellcolor{green!25}84.3          & 79.3 & 91.3                                                                       & \cellcolor{blue!25}93.4   \\
CW ($c=0.01$)   & 0             & 63.5          & \cellcolor{green!25}68.5 & 73.7                                                                       &\cellcolor{blue!25} 92.4     \\
CW ($c=0.1$)    & 0             & 41.1          & \cellcolor{green!25}63.7 & 60.5                                                                       &\cellcolor{blue!25}  81.2 \\
FGSM($\epsilon=0.02$)  & 23.5          & \cellcolor{green!25}72.5          & 70.6      & \cellcolor{blue!25}78.5                                                                       & 76.1        \\
FGSM($\epsilon=0.04$)  & 16.3          & 56.3          & \cellcolor{green!25}63.3      & 69.9                                                                       &\cellcolor{blue!25} 70.3        \\
BIM($\epsilon=0.01$)   & 0             & 62.9          & \cellcolor{green!25}68.4      & 74.5                                                                       & \cellcolor{blue!25}77.2        \\
BIM($\epsilon=0.02$)   & 0             & 40.1          & \cellcolor{green!25}53.4      & \cellcolor{blue!25}57.3                                                                       & 54.8        \\
PGD  ($\epsilon=0.01$) & 0             & 60.1          &\cellcolor{green!25} 68.8      & 75.7                                                                       &\cellcolor{blue!25} 85          \\
PGD ($\epsilon=0.02$)  & 0             & 39.3          & \cellcolor{green!25}55.2      & 58.5                                                                       &\cellcolor{blue!25} 60.2       \\ \hline
\end{tabular}
\end{table}

\begin{table}[!h]
\centering

\caption{CIFAR-100 adversarial robustness ($\%$) under the white-box threat model.}
\label{tab:cifar100_white-box}
\begin{tabular}{cccccc}
\hline
Attack       & CE  & Mustafa et al. & MadNet    &\begin{tabular}[c]{@{}c@{}}Mustafa et al.\\ Adv training\end{tabular} & \begin{tabular}[c]{@{}c@{}} MadNet \\ Adv training\end{tabular} \\ \hline
No-Attack     & 72.3          & 71.9          & 71.1   & 68.3                                                                 & 66.7        \\
BIM($\epsilon=0.005$)  & 4             & 44.8          & \cellcolor{green!25}46.8   & 55.7                                                                 & \cellcolor{blue!25}78.1        \\
BIM($\epsilon=0.01$)   & 1             & \cellcolor{green!25}39.8          & 31.2   & 46.9                                                                 & \cellcolor{blue!25}62.8        \\
PGD ($\epsilon=0.005$) & 8             & 42.2          & \cellcolor{green!25}53.4   & 55                                                                   & \cellcolor{blue!25}76          \\
PGD ($\epsilon=0.01$)  & 1             & \cellcolor{green!25}38.9          & 33.5   & 44                                                                   & \cellcolor{blue!25}63.1        \\ \hline
\end{tabular}
\end{table}

% The profound effect of adding adversarial training to MadNet, even for attacks not included in the adversarial training process, 
% can be explained by the very intuition behind the MAD loss.
% The goal of the MAD loss is to increase the margin between class cluster centroids. 
% Having an adversarial example introduced during training will be greatly penalized as it will constitute a sample that lies far from its class centroid, increasing the class variance, and very close to a different class centroid, decreasing the between-class distance.
% These attributes can potentially make MadNet even more robust if additional attacks are included during training.

\subsection{Black-Box Study}
To evaluate our model in the black-box threat model, where the attacker cannot access the model's weights, we followed \cite{papernot2017practical,carlini2019evaluating} and created a 
proxy model, which was trained using input-output pairs
probed from the defender's (target) model (i.e., the proxy model
was trained via teacher-student distillation of the target model). 
The proxy model was then used by the attacker to generate adversarial examples under the \emph{white-box} threat model. These adversarial examples were then tested against the target model. 
The proxy model we chose was ResNet-32.
One main purpose of the black-box threat model is to check for obfuscated gradients. In their paper, \citet{athalye2018obfuscated} listed defense methods that mask or obstruct the DNN's gradients.
As such, when the attacking algorithms attempts to create an adversarial example it fails to do so, simply because of its inability to properly derive the DNN's output w.r.t its input. 
This might give a false sense of security  since with slight modifications of the attacks, these defenses can be overcome.
If a defense method performs better under the white-box threat model, this would indicate an obfuscation of gradients.
The adversarial robustness results under the black-box threat model appear in Table \ref{tab:black_box}. As expected, the black-box threat model's performance is superior to that of the white-box threat model achieving, for example, $97.6$\%, $95.7$\% and $78.3$\% against the black-box PGD attack on MNIST, CIFAR-10 and CIFAR-100 respectively vs $92.4$\%, $68.8$ and $33.5$\% against its white-box counterpart.
While this alone does not guarantee the gradients are not masked or manipulated, we argue that the components that constitute the MAD loss do not impose any constraints on the model's gradients other than the derivation for the Jacobian reduction,
a process that also occurs when performing standard adversarial training.

\begin{table}[!]
\centering
\caption{Adversarial robustness ($\%$) for MadNet under the black-box threat model.}
\label{tab:black_box}
\begin{tabular}{cccc}
\hline
Attack & \begin{tabular}[c]{@{}c@{}}MNIST\\ $\epsilon=0.1$ , $c=0.1$\end{tabular} & \begin{tabular}[c]{@{}c@{}}CIFAR-10\\ $\epsilon=0.02$ , $c = 0.001$\end{tabular} & \begin{tabular}[c]{@{}c@{}}CIFAR-100\\ $\epsilon=0.01$\end{tabular} \\ \hline
\multicolumn{4}{c}{\textbf{CE}}                \\ \hline
FGSM   & 74.2  & 51.3     & 61.5         \\
BIM    & 81.7  & 49.1     & 44.3      \\
PGD    & 78.3  & 57.8     & 47.2      \\
C\&W   & 94.3  & 92.3     & 95.2         \\ \hline
\multicolumn{4}{c}{\textbf{MAD}}               \\ \hline
FGSM   & 98.1  & 85.1     & 80.8          \\
BIM    & 98    & 96.3     & 79.8      \\
PGD    & 97.6  & 95.7     & 78.3      \\
C\&W   & 100   & 100      & 98          \\ \hline
\end{tabular}
\end{table}

% \begin{table}[!h]
% \centering
% \caption{Adversarial robustness ($\%$) for MadNet under the black-box threat model.}
% \label{tab:black_box}
% \begin{tabular}{cccc}
% \hline
% Attack & \begin{tabular}[c]{@{}c@{}}MNIST\\ $\epsilon=0.1$ , $c=0.1$\end{tabular} & \begin{tabular}[c]{@{}c@{}}CIFAR-10\\ $\epsilon=0.02$ , $c = 0.001$\end{tabular} & \begin{tabular}[c]{@{}c@{}}CIFAR-100\\ $\epsilon=0.01$\end{tabular} \\ \hline
% FGSM   & 98.1                                                            & 85.1                                                                    &       -                                                       \\
% BIM    & 98                                                              & 96.3                                                                    & 79.8                                                         \\
% PGD    & 97.6                                                            & 95.7                                                                    & 78.3                                                         \\
% C\&W   & 100                                                             & 100                                                                     &               -                                               \\ \hline
% \end{tabular}
% \end{table}

\section{Conclusions}
We introduced maximum adversarial distortion (MAD), a powerful approach for the defense of deep models against adversarial attacks by optimizing for 
sensitivity-normalized embedding separability.
We demonstrated state-of-the-art results in defending against adversarial attacks.
In addition, we provided some geometric intuition on attacks and defenses
using both Davies–Bouldin Index analysis and t-SNE visualizations.
This work raises several interesting questions. First, it would be valuable to examine other methods for margin maximization and Jacobian reduction. For example, a recent work by \cite{zhang2019recurjac}
proposed an iterative  technique to reduce the norm of the Jacobian.
In addition, it would also be interesting to explore ways to increase the margin (and reduce the Jacobian) on shallower embedding layers where 
lower-level features are formed.
Finally, our MAD approach utilized a first-order
approximation and ignored the Taylor approximation error term, which can be bounded in terms of $\norm{H_{\ell}(x)}$ and 
$\norm{\epsilon}^2$, where $H_{\ell}$ is the Hessian of the embedding
layer with respect to the input. It would be valuable to try and further robustify networks by also accounting for this error 
term.

\section*{Broader Impact}
In this paper we present a strong defense method against adversarial attacks.
As such, the paper potentially has a positive societal effect because defense
techniques  constitute our only countermeasure against malicious attempts to fool deep learning systems that help or serve the public. 
We are unable to identify negative consequences of the proposed algorithms.

\bibliographystyle{abbrvnat}
\bibliography{bib}

\begin{thebibliography}{45}
\providecommand{\natexlab}[1]{#1}
\providecommand{\url}[1]{\texttt{#1}}
\expandafter\ifx\csname urlstyle\endcsname\relax
  \providecommand{\doi}[1]{doi: #1}\else
  \providecommand{\doi}{doi: \begingroup \urlstyle{rm}\Url}\fi

\bibitem[Athalye et~al.(2018)Athalye, Carlini, and
  Wagner]{athalye2018obfuscated}
A.~Athalye, N.~Carlini, and D.~Wagner.
\newblock Obfuscated gradients give a false sense of security: Circumventing
  defenses to adversarial examples.
\newblock \emph{arXiv preprint arXiv:1802.00420}, 2018.

\bibitem[Bromley et~al.(1994)Bromley, Guyon, LeCun, S{\"a}ckinger, and
  Shah]{bromley1994signature}
J.~Bromley, I.~Guyon, Y.~LeCun, E.~S{\"a}ckinger, and R.~Shah.
\newblock Signature verification using a" siamese" time delay neural network.
\newblock In \emph{Advances in neural information processing systems}, pages
  737--744, 1994.

\bibitem[Carlini and Wagner(2017{\natexlab{a}})]{carlini2017adversarial}
N.~Carlini and D.~Wagner.
\newblock Adversarial examples are not easily detected: Bypassing ten detection
  methods.
\newblock In \emph{Proceedings of the 10th ACM Workshop on Artificial
  Intelligence and Security}, pages 3--14. ACM, 2017{\natexlab{a}}.

\bibitem[Carlini and Wagner(2017{\natexlab{b}})]{carlini2017towards}
N.~Carlini and D.~Wagner.
\newblock Towards evaluating the robustness of neural networks.
\newblock In \emph{2017 IEEE Symposium on Security and Privacy (SP)}, pages
  39--57. IEEE, 2017{\natexlab{b}}.

\bibitem[Carlini et~al.(2019)Carlini, Athalye, Papernot, Brendel, Rauber,
  Tsipras, Goodfellow, and Madry]{carlini2019evaluating}
N.~Carlini, A.~Athalye, N.~Papernot, W.~Brendel, J.~Rauber, D.~Tsipras,
  I.~Goodfellow, and A.~Madry.
\newblock On evaluating adversarial robustness.
\newblock \emph{arXiv preprint arXiv:1902.06705}, 2019.

\bibitem[Chan et~al.(2019)Chan, Tay, Ong, and Fu]{chan2019jacobian}
A.~Chan, Y.~Tay, Y.~S. Ong, and J.~Fu.
\newblock Jacobian adversarially regularized networks for robustness.
\newblock \emph{arXiv preprint arXiv:1912.10185}, 2019.

\bibitem[Cohen et~al.(2019)Cohen, Rosenfeld, and Kolter]{cohen2019certified}
J.~M. Cohen, E.~Rosenfeld, and J.~Z. Kolter.
\newblock Certified adversarial robustness via randomized smoothing.
\newblock \emph{arXiv preprint arXiv:1902.02918}, 2019.

\bibitem[Dathathri et~al.(2018)Dathathri, Zheng, Murray, and
  Yue]{dathathri2018detecting}
S.~Dathathri, S.~Zheng, R.~M. Murray, and Y.~Yue.
\newblock Detecting adversarial examples via neural fingerprinting.
\newblock \emph{arXiv preprint arXiv:1803.03870}, 2018.

\bibitem[Davies and Bouldin(1979)]{davies1979cluster}
D.~L. Davies and D.~W. Bouldin.
\newblock A cluster separation measure.
\newblock \emph{IEEE transactions on pattern analysis and machine
  intelligence}, \penalty0 (2):\penalty0 224--227, 1979.

\bibitem[Dhillon et~al.(2018)Dhillon, Azizzadenesheli, Lipton, Bernstein,
  Kossaifi, Khanna, and Anandkumar]{dhillon2018stochastic}
G.~S. Dhillon, K.~Azizzadenesheli, Z.~C. Lipton, J.~Bernstein, J.~Kossaifi,
  A.~Khanna, and A.~Anandkumar.
\newblock Stochastic activation pruning for robust adversarial defense.
\newblock \emph{arXiv preprint arXiv:1803.01442}, 2018.

\bibitem[Ding et~al.(2018)Ding, Sharma, Lui, and Huang]{ding2018max}
G.~W. Ding, Y.~Sharma, K.~Y.~C. Lui, and R.~Huang.
\newblock Max-margin adversarial (mma) training: Direct input space margin
  maximization through adversarial training.
\newblock \emph{arXiv preprint arXiv:1812.02637}, 2018.

\bibitem[Elsayed et~al.(2018)Elsayed, Krishnan, Mobahi, Regan, and
  Bengio]{elsayed2018large}
G.~Elsayed, D.~Krishnan, H.~Mobahi, K.~Regan, and S.~Bengio.
\newblock Large margin deep networks for classification.
\newblock In \emph{Advances in neural information processing systems}, pages
  842--852, 2018.

\bibitem[Goodfellow et~al.(2014)Goodfellow, Shlens, and
  Szegedy]{goodfellow2014explaining}
I.~J. Goodfellow, J.~Shlens, and C.~Szegedy.
\newblock Explaining and harnessing adversarial examples.
\newblock \emph{arXiv preprint arXiv:1412.6572}, 2014.

\bibitem[He et~al.(2016)He, Zhang, Ren, and Sun]{he2016deep}
K.~He, X.~Zhang, S.~Ren, and J.~Sun.
\newblock Deep residual learning for image recognition.
\newblock In \emph{Proceedings of the IEEE conference on computer vision and
  pattern recognition}, pages 770--778, 2016.

\bibitem[Hein and Andriushchenko(2017)]{hein2017formal}
M.~Hein and M.~Andriushchenko.
\newblock Formal guarantees on the robustness of a classifier against
  adversarial manipulation.
\newblock In \emph{Advances in Neural Information Processing Systems}, pages
  2266--2276, 2017.

\bibitem[Hoffer and Ailon(2015)]{hoffer2015deep}
E.~Hoffer and N.~Ailon.
\newblock Deep metric learning using triplet network.
\newblock In \emph{International Workshop on Similarity-Based Pattern
  Recognition}, pages 84--92. Springer, 2015.

\bibitem[Krizhevsky and Hinton(2009)]{krizhevsky2009learning}
A.~Krizhevsky and G.~Hinton.
\newblock Learning multiple layers of features from tiny images.
\newblock Technical report, Citeseer, 2009.

\bibitem[Kurakin et~al.(2016{\natexlab{a}})Kurakin, Goodfellow, and
  Bengio]{kurakin2016adversarial}
A.~Kurakin, I.~Goodfellow, and S.~Bengio.
\newblock Adversarial examples in the physical world.
\newblock \emph{arXiv preprint arXiv:1607.02533}, 2016{\natexlab{a}}.

\bibitem[Kurakin et~al.(2016{\natexlab{b}})Kurakin, Goodfellow, and
  Bengio]{kurakin2016adversarial2}
A.~Kurakin, I.~Goodfellow, and S.~Bengio.
\newblock Adversarial machine learning at scale.
\newblock \emph{arXiv preprint arXiv:1611.01236}, 2016{\natexlab{b}}.

\bibitem[LeCun et~al.(1998)LeCun, Bottou, Bengio, Haffner,
  et~al.]{lecun1998gradient}
Y.~LeCun, L.~Bottou, Y.~Bengio, P.~Haffner, et~al.
\newblock Gradient-based learning applied to document recognition.
\newblock \emph{Proceedings of the IEEE}, 86\penalty0 (11):\penalty0
  2278--2324, 1998.

\bibitem[Liang et~al.(2017)Liang, Wang, Lei, Liao, and Li]{liang2017soft}
X.~Liang, X.~Wang, Z.~Lei, S.~Liao, and S.~Z. Li.
\newblock Soft-margin softmax for deep classification.
\newblock In \emph{International Conference on Neural Information Processing},
  pages 413--421. Springer, 2017.

\bibitem[Liu et~al.(2016)Liu, Wen, Yu, and Yang]{liu2016large}
W.~Liu, Y.~Wen, Z.~Yu, and M.~Yang.
\newblock Large-margin softmax loss for convolutional neural networks.
\newblock In \emph{ICML}, volume~2, page~7, 2016.

\bibitem[Maaten and Hinton(2008)]{maaten2008visualizing}
L.~v.~d. Maaten and G.~Hinton.
\newblock Visualizing data using t-sne.
\newblock \emph{Journal of machine learning research}, 9\penalty0
  (Nov):\penalty0 2579--2605, 2008.

\bibitem[Madry et~al.(2017)Madry, Makelov, Schmidt, Tsipras, and
  Vladu]{madry2017towards}
A.~Madry, A.~Makelov, L.~Schmidt, D.~Tsipras, and A.~Vladu.
\newblock Towards deep learning models resistant to adversarial attacks.
\newblock \emph{arXiv preprint arXiv:1706.06083}, 2017.

\bibitem[M{\"u}ller et~al.(2019)M{\"u}ller, Kornblith, and
  Hinton]{muller2019does}
R.~M{\"u}ller, S.~Kornblith, and G.~Hinton.
\newblock When does label smoothing help?
\newblock \emph{arXiv preprint arXiv:1906.02629}, 2019.

\bibitem[Mustafa et~al.(2019)Mustafa, Khan, Hayat, Goecke, Shen, and
  Shao]{mustafa2019adversarial}
A.~Mustafa, S.~Khan, M.~Hayat, R.~Goecke, J.~Shen, and L.~Shao.
\newblock Adversarial defense by restricting the hidden space of deep neural
  networks.
\newblock In \emph{Proceedings of the IEEE International Conference on Computer
  Vision}, pages 3385--3394, 2019.

\bibitem[Pang et~al.(2019)Pang, Xu, Du, Chen, and Zhu]{pang2019improving}
T.~Pang, K.~Xu, C.~Du, N.~Chen, and J.~Zhu.
\newblock Improving adversarial robustness via promoting ensemble diversity.
\newblock \emph{arXiv preprint arXiv:1901.08846}, 2019.

\bibitem[Papernot et~al.(2016)Papernot, McDaniel, Wu, Jha, and
  Swami]{papernot2016distillation}
N.~Papernot, P.~McDaniel, X.~Wu, S.~Jha, and A.~Swami.
\newblock Distillation as a defense to adversarial perturbations against deep
  neural networks.
\newblock In \emph{2016 IEEE Symposium on Security and Privacy (SP)}, pages
  582--597. IEEE, 2016.

\bibitem[Papernot et~al.(2017)Papernot, McDaniel, Goodfellow, Jha, Celik, and
  Swami]{papernot2017practical}
N.~Papernot, P.~McDaniel, I.~Goodfellow, S.~Jha, Z.~B. Celik, and A.~Swami.
\newblock Practical black-box attacks against machine learning.
\newblock In \emph{Proceedings of the 2017 ACM on Asia conference on computer
  and communications security}, pages 506--519. ACM, 2017.

\bibitem[Ross and Doshi-Velez(2018)]{ross2018improving}
A.~S. Ross and F.~Doshi-Velez.
\newblock Improving the adversarial robustness and interpretability of deep
  neural networks by regularizing their input gradients.
\newblock In \emph{Thirty-second AAAI conference on artificial intelligence},
  2018.

\bibitem[Sokoli{\'c} et~al.(2017)Sokoli{\'c}, Giryes, Sapiro, and
  Rodrigues]{sokolic2017robust}
J.~Sokoli{\'c}, R.~Giryes, G.~Sapiro, and M.~R. Rodrigues.
\newblock Robust large margin deep neural networks.
\newblock \emph{IEEE Transactions on Signal Processing}, 65\penalty0
  (16):\penalty0 4265--4280, 2017.

\bibitem[Strauss et~al.(2017)Strauss, Hanselmann, Junginger, and
  Ulmer]{strauss2017ensemble}
T.~Strauss, M.~Hanselmann, A.~Junginger, and H.~Ulmer.
\newblock Ensemble methods as a defense to adversarial perturbations against
  deep neural networks.
\newblock \emph{arXiv preprint arXiv:1709.03423}, 2017.

\bibitem[Sun et~al.(2016)Sun, Chen, Wang, Liu, and Liu]{sun2016depth}
S.~Sun, W.~Chen, L.~Wang, X.~Liu, and T.-Y. Liu.
\newblock On the depth of deep neural networks: A theoretical view.
\newblock In \emph{Thirtieth AAAI Conference on Artificial Intelligence}, 2016.

\bibitem[Szegedy et~al.(2013)Szegedy, Zaremba, Sutskever, Bruna, Erhan,
  Goodfellow, and Fergus]{szegedy2013intriguing}
C.~Szegedy, W.~Zaremba, I.~Sutskever, J.~Bruna, D.~Erhan, I.~Goodfellow, and
  R.~Fergus.
\newblock Intriguing properties of neural networks.
\newblock \emph{arXiv preprint arXiv:1312.6199}, 2013.

\bibitem[Szegedy et~al.(2016)Szegedy, Vanhoucke, Ioffe, Shlens, and
  Wojna]{szegedy2016rethinking}
C.~Szegedy, V.~Vanhoucke, S.~Ioffe, J.~Shlens, and Z.~Wojna.
\newblock Rethinking the inception architecture for computer vision.
\newblock In \emph{Proceedings of the IEEE conference on computer vision and
  pattern recognition}, pages 2818--2826, 2016.

\bibitem[Tjeng et~al.(2018)Tjeng, Xiao, and Tedrake]{tjeng2018evaluating}
V.~Tjeng, K.~Y. Xiao, and R.~Tedrake.
\newblock Evaluating robustness of neural networks with mixed integer
  programming.
\newblock 2018.

\bibitem[Tram{\`e}r et~al.(2017)Tram{\`e}r, Kurakin, Papernot, Goodfellow,
  Boneh, and McDaniel]{tramer2017ensemble}
F.~Tram{\`e}r, A.~Kurakin, N.~Papernot, I.~Goodfellow, D.~Boneh, and
  P.~McDaniel.
\newblock Ensemble adversarial training: Attacks and defenses.
\newblock \emph{arXiv preprint arXiv:1705.07204}, 2017.

\bibitem[Tsuzuku et~al.(2018)Tsuzuku, Sato, and Sugiyama]{tsuzuku2018lipschitz}
Y.~Tsuzuku, I.~Sato, and M.~Sugiyama.
\newblock Lipschitz-margin training: Scalable certification of perturbation
  invariance for deep neural networks.
\newblock In \emph{Advances in Neural Information Processing Systems}, pages
  6541--6550, 2018.

\bibitem[Wang et~al.(2018)Wang, Wang, Zhou, Ji, Gong, Zhou, Li, and
  Liu]{wang2018cosface}
H.~Wang, Y.~Wang, Z.~Zhou, X.~Ji, D.~Gong, J.~Zhou, Z.~Li, and W.~Liu.
\newblock Cosface: Large margin cosine loss for deep face recognition.
\newblock In \emph{Proceedings of the IEEE Conference on Computer Vision and
  Pattern Recognition}, pages 5265--5274, 2018.

\bibitem[Wen et~al.(2016)Wen, Zhang, Li, and Qiao]{wen2016discriminative}
Y.~Wen, K.~Zhang, Z.~Li, and Y.~Qiao.
\newblock A discriminative feature learning approach for deep face recognition.
\newblock In \emph{European conference on computer vision}, pages 499--515.
  Springer, 2016.

\bibitem[Wong and Kolter(2017)]{wong2017provable}
E.~Wong and J.~Z. Kolter.
\newblock Provable defenses against adversarial examples via the convex outer
  adversarial polytope.
\newblock \emph{arXiv preprint arXiv:1711.00851}, 2017.

\bibitem[Wong et~al.(2018)Wong, Schmidt, Metzen, and Kolter]{wong2018scaling}
E.~Wong, F.~Schmidt, J.~H. Metzen, and J.~Z. Kolter.
\newblock Scaling provable adversarial defenses.
\newblock In \emph{Advances in Neural Information Processing Systems}, pages
  8400--8409, 2018.

\bibitem[Yan et~al.(2018)Yan, Guo, and Zhang]{yan2018deep}
Z.~Yan, Y.~Guo, and C.~Zhang.
\newblock Deep defense: Training dnns with improved adversarial robustness.
\newblock In \emph{Advances in Neural Information Processing Systems}, pages
  417--426, 2018.

\bibitem[Yu et~al.(2018)Yu, Liu, Wang, Zhao, and Chen]{yu2018interpreting}
F.~Yu, C.~Liu, Y.~Wang, L.~Zhao, and X.~Chen.
\newblock Interpreting adversarial robustness: A view from decision surface in
  input space.
\newblock \emph{arXiv preprint arXiv:1810.00144}, 2018.

\bibitem[Zhang et~al.(2019)Zhang, Zhang, and Hsieh]{zhang2019recurjac}
H.~Zhang, P.~Zhang, and C.-J. Hsieh.
\newblock Recurjac: An efficient recursive algorithm for bounding jacobian
  matrix of neural networks and its applications.
\newblock In \emph{Proceedings of the AAAI Conference on Artificial
  Intelligence}, volume~33, pages 5757--5764, 2019.

\end{thebibliography}

\newpage
\clearpage
\begin{appendices}
  %\chapter{Appendix}
  \section{$L_2$ Norm Sub-Multiplicativity}
  \label{appendix_sec:Frobenius}

  Given $A \in \R^{N\times M \times K}$ and $B \in \R^{K \times J \times L}$, we claim that the Frobenius norm of the multiplication of $A$ and $B$ is less than or equal to the multiplication of each tensor's Frobenius norm. Proof:

 \begin{align}
\|AB\|^2_F&=\sum\limits_{n=1}^{N}\sum\limits_{m=1}^{M}\sum\limits_{j=1}^{J}\sum\limits_{l=1}^{L}\left|\sum\limits_{k=1}^Ka_{n,m,k}b_{k,j,l}\right|^2
\nonumber
\\
&\leqslant\sum\limits_{n=1}^{N}\sum\limits_{m=1}^{M}\sum\limits_{j=1}^{J}\sum\limits_{l=1}^{L}\sum\limits_{k=1}^K\left|a_{n,m,k}\right|^2\sum\limits_{k=1}^K\left|b_{k,j,l}\right|^2 \tag{Cauchy-Schwarz}
\nonumber
\\
&=\sum\limits_{n=1}^{N}\sum\limits_{m=1}^{M}\sum\limits_{j=1}^{J}\sum\limits_{l=1}^{L}(\sum\limits_{k,s=1}^K\left|a_{n,m,k}\right|^2\left|b_{k,j,l}\right|^2)
\nonumber
\\
&=\sum\limits_{n=1}^{N}\sum\limits_{m=1}^{M}\sum\limits_{k=1}^K\left|a_{n,m,k}\right|^2\sum\limits_{j=1}^{J}\sum\limits_{l=1}^{L}\sum\limits_{s=1}^K\left|b_{s,j,l}\right|^2
\\\nonumber
&=\|A\|^2_F\|B\|^2_F
\end{align}

\section{MAD Algorithm}
The MAD algorithm appears in Algorithm \ref{alg:psudo}.
 \label{appendix_sec:mad}
 
 \begin{algorithm}[!htb]
 \centering
\caption{MadNet optimization} \label{alg:Mad_Training_Process}
\begin{algorithmic}[1]
\Procedure{MAD}{}

 \For{batch $= 1,\ldots,$\#batches}
 
    \State $X,Y \gets$ get\_batch()
    \State initialize $X_{\text{siamese}} = [] , Y_{\text{siamese}} = [] , S = []$
    \For{$b=1,\ldots,\text{batch\_size} $}
    \State $q \sim \text{Bernoulli}(Q)$
    \If{$q==1$}
        \State $y \gets Y[b] $ 
        \State $s \gets 1$
  \Else   
        \State $y  \gets$ random class $\neq y_1$
        \State $s \gets 0$ 
    \EndIf
    \State $x \gets $ random sample from class $Y[b]$  
    \State Append $x,y,s$ to $X_{\text{siamese}}, Y_{\text{siamese}} , S $
  \EndFor
     \State $z_1,z_2 \gets  F_{\ell}(X),F_{\ell}{X_{\text{siamese}})}$  \Comment{embedding}
  \State $p_1,p_2 \gets  F(X),F(X_{\text{siamese}})$\Comment{logits}
    \State $\text{SL} = \frac{1}{\text{batch\_size}}\sum|\frac{z_1z_2}{\norm{z_1}\norm{z_2}}-S| $ \
 
 \State $RVL \gets 0$ 
    \State $JL \gets \norm{\frac{\partial F(x_1)}{\partial x_1}} +\norm{\frac{\partial F(x_2)}{\partial x_2}} $ 

  \For{$c=0,\ldots,M$}
  \State $\mu_c = \frac{1}{N_c}\sum_{i=1}^{N_c}{z^c_i}$ 
  \State $\sigma_c = \frac{1}{N_c}\sum_{i=1}^{N_c}{\norm{z^c_i - \mu_c}_2} $ 
  \State $\text{RVL} \gets \text{RVL} + \sigma_c$ 
\EndFor
\State  \text{minimize}   ${\cal L}_{MAD}$
 \EndFor

\EndProcedure
\end{algorithmic}
\label{alg:psudo}
\end{algorithm}

\section{Classifier Hyperparameters}
\label{appendix_sec:classifier_params}

\begin{table}[!htb]
\centering
    \begin{tabular}{cc}
    \hline
Parameter             & Value            \\ \hline
Optimizer             & SGD              \\
ResNet Depth          & 56               \\
Weight Regularization & L2 (0.002)       \\
Batch Size            & 128              \\
Initial Learning Rate & 0.1              \\
Epochs-CIFAR-10        & 400              \\
Epochs-MNIST          & 80               \\
Activation            & Leaky-Relu (0.1) \\
$\lambda_{\text{CE}}$             & 1                 \\
$\lambda_{\text{siam}}$     & 1                 \\
$\lambda_{\text{RVL}}$         & 1                 \\
$\lambda_{\text{Jacobian}}$   & 0.01          \\
$\alpha$                & 0.8          \\
\hline
    \end{tabular}
\end{table}

\section{Adversarial Attacks}
\label{appendix_sec:adversarial_attacks}
We use various attack algorithms to evaluate our defense method.
We denote $x$,$x'$ as the input and adversarial instance, respectively,
$\ell$ as the target label,
$F$ as the target model with loss function $L_F(x,\ell)$
and $\epsilon=||x-x'||$ as the pixel-wise perturbation between the adversarial and normal instances.
The general formulation, therefore, becomes,
\begin{equation}
    \underset{x'}{\text{minimize}} \ ||x-x'||^2 \ \  s.t.\  F(x')= \ell.
    \label{eq:adversarial_optimization_problem}
\end{equation}

\textbf{FGSM}: \citet{goodfellow2014explaining} introduced the fast gradient sign method (FGSM), which optimizes the  adversarial instance by back-propagating the input through the attacked DNN, in accordance with the desired target. Formally, letting $\epsilon$ be a fixed parameter, 
the adversarial example is,
\begin{equation}
x' = x + \epsilon \sign(\nabla L_F(x,\ell).
\label{eq:fgsm}
\end{equation}
While not as effective as other attack algorithms, this method has the advantage of being one of the fastest ones.

\textbf{BIM}:
\citet{kurakin2016adversarial} introduced the Basic Iterative Method (BIM), which performs the FGSM method iteratively, clipping the perturbation if needed. Formally,
$$
x'_{N+1} = x'_N + \epsilon \sign(\nabla L_F(x'_N,l)),
$$
where $\epsilon$ is a fixed parameter.

 \textbf{PGD}:
 \citet{madry2017towards} proposed an attack similar to BIM, performing FGSM steps from a sample randomly selected from a neighbourhood of $x$ .
 
\textbf{C\&W}: \citet{carlini2017towards} introduced the C\&W method,
which operates by modifying the L-BFGS method as follows,
$$
\underset{x'}{\text{minimize}} \  ||x-x'||^2 + cf(x'),
$$
where $c$ is a hyperparameter and the loss function,
$f$, is chosen such that $f(x') <= 0$ if $x'$ is classified 
as the target class; namely,
$$
    f(x') = \max(Z(x')_{\ell}-Z(x')_t,\kappa)
$$
where $Z$ is the target DNN logits, $t$ is the correct label and $\kappa$ is a hyperparameter referring to the confidence of the attack.
The higher the confidence, the higher the activation for the target class
is, and therefore, the larger the perturbation.
Three different Euclidean norms are considered with this algorithm, $L_0, L_2, L_\infty$. 
We conduct our evaluation using
$L_2$.
This attack method is considered very effective and has had great success in overcoming various defense methods \cite{papernot2016distillation,carlini2017towards}.

\section{Adversarial Attack Parameters}
\label{appendix_sec:adversarial_params}
The parameters chosen to create the adversarial examples per attack are given in Table 
\ref{tab:adversarial_params}.

\begin{table}[!htb]
\centering
      \caption{Adversarial attack parameters. }

\begin{tabular}{ccc}
 \hline
Attack & Parameter           & Value      \\ \hline
BIM    & Iterations          & 10   \\ 
PGD    & Iterations          & 10   \\ \hline
CW     & Max Iterations      & 1000         \\
      & Binary Search Steps & 1          \\
      & Confidence          & 0
     \\\hline
\end{tabular}
\label{tab:adversarial_params}
\end{table}

\section{Label Smoothing}
\label{appendix_sec:smoothing}
 Label smoothing, the process of training a DNN with soft labels rather than a one-hot vector,  has been shown to decrease the gradients w.r.t. the model's weights \cite{muller2019does}, while not directly used in the context of adversarial robustness.
 Intuitively, the ``cost'' of an incorrect prediction is lower when the target label's activation is lower than $1$.
 As such, the update increment of the weights is lower than in standard one-hot training.
 By the same logic, if we consider the weights fixed and the input instance as being updated according to a gradient step, as is the basis of adversarial attacks, we can deduce that the gradients will be lower than those of a standard training procedure.
 We set the softmax response of the correct class to $\alpha$ and the remaining classes are distributed evenly to sum to $1$.
 
%  \section{resubmission comments}
%  Context and details of resubmission. Note: this information will be shared only with ACs and SACs, and only after the initial desk rejection phase.
% If your work is a resubmission, describe:
% - the venue and year where this work was submitted
% - a summary of the reviews and the reasons given for rejection (if any)
% - the changes that have been made to the paper since the last submission.

% The paper was submitted to ICML 2020 and withdrawn during rebuttal. 
% Previous version focused on adversarial detection based on density estimation of latent spaces (Feinman et al 2017.). Reviewers claimed that such approach is obsolete and emphasized the importance of robustness over detection in the adversarial defense domain.
% We decided to follow this advice and redesigned the 
% algorithm to focus on robustness. 
% More importantly, we significantly expanded the scope of our empirical study, including CIFAR-100 and comparing to many more baselines and threat models. 
% Overall the results we present in this version are by far more convincing, demonstrate the ability to scale up the algorithm to larger datasets, and now strictly advance the state-of-the-art in terms of robustness performance over a wide variety of attacks and threat models.
\end{appendices}

\end{document}